\documentclass[runningheads]{llncs}

\usepackage{eccv}

\usepackage{authblk}

\usepackage{eccvabbrv}

\usepackage{graphicx}
\usepackage{booktabs}

\usepackage{algorithm}
\usepackage{algorithmic}
\usepackage{amsmath}

\usepackage{makecell}
\usepackage{multirow}
\graphicspath{{figs/}}
\usepackage{tabularx}
\usepackage{bbding}
\usepackage{pifont}
\usepackage{bm}
\usepackage{wrapfig}

\newcommand{\op}[1]{\operatorname{#1}}

\usepackage[accsupp]{axessibility}  

\usepackage[pagebackref,breaklinks,colorlinks,citecolor=eccvblue]{hyperref}

\usepackage{orcidlink}

\begin{document}

\title{Fourier or Wavelet bases as counterpart self-attention in spikformer for efficient visual classification} 

\titlerunning{Abbreviated paper title}

\author{Qingyu Wang\inst{1,2*} \and
Duzhen Zhang\inst{1,2*} \and
Tilelin Zhang\inst{1,2\dagger} \and
Bo Xu\inst{1,2}}



\authorrunning{Qingyu et al.}

\institute{School of Artificial Intelligence, University of Chinese Academy of Sciences \and
Institute of Automation, Chinese Academy of Sciences\\
\email{\{wangqingyu2022, zhangduzhen2019, tielin.zhang, xubo\}@ia.ac.cn}}

\maketitle

\begin{abstract}
  Energy-efficient spikformer has been proposed by integrating the biologically plausible spiking neural network (SNN) and artificial Transformer, whereby the Spiking Self-Attention (SSA) is used to achieve both higher accuracy and lower computational cost. However, it seems that self-attention is not always necessary, especially in sparse spike-form calculation manners. In this paper, we innovatively replace vanilla SSA (using dynamic bases calculating from Query and Key) with spike-form Fourier Transform, Wavelet Transform, and their combinations (using fixed triangular or wavelets bases), based on a key hypothesis that both of them use a set of basis functions for information transformation. Hence, the \textbf{F}ourier-or-\textbf{W}avelet-based spikformer (\textbf{FWformer}) is proposed and verified in visual classification tasks, including both static image and event-based video datasets. The FWformer can achieve comparable or even higher accuracies ($0.4\%$-$1.5\%$), higher running speed ($9\%$-$51\%$ for training and $19\%$-$70\%$ for inference), reduced theoretical energy consumption ($20\%$-$25\%$), and reduced GPU memory usage ($4\%$-$26\%$), compared to the standard spikformer. Our result indicates the continuous refinement of new Transformers, that are inspired either by biological discovery (spike-form), or information theory (Fourier or Wavelet Transform), is promising.
  \keywords{spiking neural network \and Transformer \and visual classification}
\end{abstract}

\section{Introduction}
\label{sec:intro}

Spiking neural network (SNN) is considered the third generation of artificial neural networks~\cite{maass1997networks} for its biological plausibility of event-driven characteristics. It has also received extensive attention in the computation area of neuromorphic hardware~\cite{davies2018loihi}, exhibiting a remarked lower computational cost on various machine learning tasks, including but not limited to, visual classification ~\cite{zhou2022spikformer}, temporal auditory recognition ~\cite{wang2023complex}, and reinforcement learning ~\cite{tang2021deep}.
The progress in SNN is contributed initially by some key computational modules inspired by the biological brain, e.g., the receptive-field-like convolutional circuits, self-organized plasticity propagation~\cite{RN767}, and other multi-scale inspiration from the single neuron or synapse to the network or cognitive functions. Simultaneously, the SNN also learns from the artificial neural network (ANN) by borrowing some mathematical optimization algorithms, e.g., the approximate gradients in backpropagation (BP), various types of loss definitions, and regression configurations.

Even though various advanced architectures have been proposed and contributed ANN to a powerful framework, the efforts to promote its training speed and computational consumption have never been stopped. As the well-known Transformer for example, it contains a rich information representation formed by multi-head self-attention, which calculates Query, Key, and Value from the inputs to connect each token in a sequence with every other token. Although having achieved rapid and widespread application, the $\mathcal{O}(N^2)$ complexity (with $N$ representing the sequence length) results in a huge training cost in Transformer that can not be neglected. Many works have tried to solve this problem, including but not limited to, replacing self-attention with unparameterized transform formats, for example, using Fourier Transform (FNet~\cite{FNET}) or Gaussian Transform (Gaussian attention~\cite{you2020hard}). Another attempt is to integrate some key features of ANNs and SNNs to exhibit their advantages, such as the higher accuracy performance in ANNs and the lower computational cost in SNNs.

The spikformer~\cite{zhou2022spikformer} explores self-attention in SNN for more advanced deep learning. It introduces a spike-form self-attention called Spiking Self-Attention (SSA). In SSA, the floating Query, Key, and Value signals are sent to leaky-integrated and fire (LIF) neurons to generate spike sequences that only contain binary and sparse $0$ and $1$ vision information, which results in non-negativeness spiking attention map. This special map doesn't require the complex softmax operation anymore for further normalization, which means a lower computational consumption is needed compared to that in vanilla Self-Attention. However, even though many efforts have been made, it seems that the SSA still exhibits an $\mathcal{O}(N^2)$ complexity, whereby further refinement is necessary. Given binary and sparse spikes for information representation, we here question whether it is still necessary to retain the original complex structure of Self-Attention in spikformer. Here, we give a hypothesis that although Self-Attention with learning parameters has been generally considered more flexible, it is still not suitable in the spike stream context, since the correlation between sparse spike trains is too weak to form closed similarity. Hence, an intuitive approach is to convert these sparse spike trains in spatial domains to the equivalent frequency domains with the help of Fourier transformation.

Here we propose a new hypothesis: Just like the Fourier Transform, Self-Attention can also be thought of as using a set of basis frequency functions for information representation. The main difference between these two methods is that the Fourier Transform uses fixed triangular basis functions to transform signals into the frequency domain, while on the contrary, the Self-Attention calculates higher-order signal representation from compositions of the input to produce more complex basis functions ($Query \times Key$). This understanding may explain why FNet~\cite{FNET} performs well, since fixed basis functions may also work in some cases by offering structured prior information. Following this perspective, an intuitive plan is to integrate all these key features together, towards a reduced computational cost and accelerated running speed, including unparameterized transforms (e.g., Fourier Transform and Wavelet Transform), and spike-form sparse representation. Our main contributions can be summarized as follows:

\begin{quotation}
    We propose a key hypothesis that the Self-Attention in Transformer works by using a set of basis functions to transform information from Query, Key, and Value sequences, which is very similar to the Fourier Transform. Hence, after jointly considering the shortcomings of spikformer, we replaced SSA with spike-form Fourier Transform and Wavelet Transform. Mathematical analysis indicates a reduced time complexity from $\mathcal{O}(Nd^2)$ or $\mathcal{O}(N^2d)$, to $\mathcal{O}(N\log N)$ or $\mathcal{O}(D\log D)+\mathcal{O}(N\log N)$, under the same accuracy performance.

    The results validate that our method achieves superior accuracy on event-based video datasets (improved by $0.3\%$-$1.2\%$) and comparable performance on spatial image datasets, compared to spikformer with SSA. Furthermore, it exhibits significantly enhanced computational efficiency, reducing memory usage by $4\%$-$26\%$, reducing theoretical energy consumption by $20\%$-$25\%$, and achieving approximately $9\%$-$51\%$ and $19\%$-$70\%$ improvements in training and inference speeds, respectively.
    
    We further analyze the orthogonality of self-attention as a set of basis functions. We find during training, that the orthogonality is continuously decreasing, which inspires us to use combined different wavelet bases with nonlinear, learnable parameters as coefficients to form structured non-orthogonal basis functions. In the second round of experiments, the experiments show even better accuracy performance on event-based video datasets (improved by $0.4\%$-$1.5\%$ compared to spikformer).
\end{quotation}

\section{Related Work}

\paragraph{Vision Transformers}
The vanilla Transformer architecture, initially designed for natural language processing~\cite{vaswani2017attention}, has then demonstrated remarkable success in various other computer-vision tasks, including image classification~\cite{dosovitskiyimage}, semantic segmentation~\cite{wang2021pyramid}, object detection~\cite{carion2020end}, and low-level image processing~\cite{chen2021pre}. The critical component that contributes to the success of the Transformer is the self-attention mechanism. In Vision Transformer (ViT), self-attention can capture global dependencies between image patches and generate meaningful representations by weighting the features of these patches, using the dot-product operation between Query and Key, followed by the softmax normalization~\cite{katharopoulos2020transformers}. The structure of ViT also fits for conventional SNNs, offering potential Transformer-type architectures for achieving higher accuracy performance.

\paragraph{Spiking Neural Networks} 

In contrast to traditional ANNs that employ continuous floating-point values to convey information, SNNs utilize discrete spike sequences for communication, offering a promising energy-efficient and biologically plausible alternative for computation. The critical components of SNNs encompass spiking neuron models, optimization algorithms, and network architectures. Spiking neurons serve as the fundamental non-linear spatial and temporal information processing units in SNNs, responsible for receiving from continuous inputs and converting them to spike sequences. Leaky Integrate-and-Fire (LIF)~\cite{dayan2005theoretical}, PLIF~\cite{fang2021incorporating}, Izhikevich~\cite{izhikevich2004spike} neurons are commonly used dynamic neuron models in SNNs for their efficiency and simplicity. There are primarily two optimization algorithms employed in deep SNNs: ANN-to-SNN conversion and direct training. 
In ANN-to-SNN conversion~\cite{rueckauer2017conversion}, a high-performance pre-trained ANN is converted into an SNN by replacing Rectified Linear Unit (ReLU) activation functions with spiking neurons. However, the converted SNN requires significant time steps to accurately approximate the ReLU activation, leading to substantial latency~\cite{han2020rmp}. In direct training, SNNs are unfolded over discrete simulation time steps and trained using backpropagation through time~\cite{shrestha2018slayer}. Since the event-triggered mechanism in spiking neurons is non-differentiable, surrogate gradients are employed to approximate the non-differentiable parts during backpropagation by using some predefined gradient values to replace infinite gradients~\cite{lee2020enabling}. 

With the advancements in ANNs, SNNs have improved their performance by incorporating advanced architectures from ANNs. These architectures include Spiking Recurrent Neural Networks~\cite{lotfi2020long}, ResNet-like SNNs~\cite{hu2021spiking}, and Spiking Graph Neural Networks~\cite{DBLP:conf/ijcai/Xu00LLP21}. Recently, exploring Transformer in the context of SNNs has received a lot of attention.
For example, temporal attention has been proposed to reduce redundant simulation time steps~\cite{yao2021temporal}. Additionally, an ANN-SNN conversion Transformer has been introduced, but it still retains vanilla self-attention that does not align with the inherent properties of SNNs~\cite{mueller2021spiking}. Furthermore, spikformer~\cite{zhou2022spikformer} investigates the feasibility of implementing self-attention and Transformer in SNNs using a direct training manner.

In this paper, we argue that the artificial Transformer can be well integrated into SNNs for higher performance, while at the same time, the utilization of SSA in spiking Transformer (spikformer) can be further replaced by a special module based on Fourier Transform or Wavelet Transform, which to some extent, indicating an alternative more efficient effort to achieve fast, efficient computation without affecting the accuracy. 

\section{Preliminaries}
\subsection{Spiking Neuron Model}
The spiking neuron serves as the fundamental unit in SNNs. It receives the current sequence and accumulates membrane potential, which is subsequently compared to a threshold to determine whether a spike should be generated. In this paper, we consistently employ LIF at all Spiking Neuron Layers.

The dynamic model of the LIF neuron is described as follows:
\begin{align}
H[t]&=V[t-1]+\frac{1}{\tau}\left(X[t]-\left(V[t-1]-V_{\text {reset }}\right)\right),\\
S[t]&=\mathcal{G}\left(H[t]-V_{t h}\right) , \\
V[t]&=H[t](1-S[t])+V_{\text {reset }} S[t],
\end{align}
where $\tau$ represents the membrane time constant, and $X[t]$ denotes the input current at time step $t$. When the membrane potential $H[t]$ exceeds the firing threshold $V_{th}$, the spiking neuron generates a spike $S[t]$. The Heaviside step function $\mathcal{G}(v)$ is defined as 1 when $v\geq 0$ and 0 otherwise. The membrane potential $V[t]$ will transition to the reset potential $V_{reset}$ if there is a spike event, or otherwise it remains unchanged as $H[t]$.

\subsection{Spiking Self-Attention}

The spikformer utilizes the SSA as its primary module for extracting sparse visual features and mixing spike sequences. Given input spike sequences denoted as $\bm{X} \in \mathbb{R}^{T \times N \times D}$, where $T$, $N$, and $D$ represent the time steps, sequence length, and feature dimension, respectively, SSA incorporates three key components: Query ($\bm{Q}$), Key ($\bm{K}$), and Value ($\bm{V}$). These components are initially obtained by applying learnable matrices $\bm{W}_Q, \bm{W}_K, \bm{W}_V \in\mathbb{R}^{D\times D }$ to the input sequences $\bm{X}$. Subsequently, they are transformed into spike sequences through Spiking Neuron Layers, formulated as:
\begin{align}
\bm{Q} = {{\mathcal{SN}}}(\op{BN}(\bm{X}\bm{W}_Q)), \bm{K}={{\mathcal{SN}}}(\op{BN}(\bm{X}\bm{W}_K)), \bm{V} = {{\mathcal{SN}}}(\op{BN}(\bm{X}\bm{W}_V)),
\label{eq:spikeqkv}
\end{align}
where $\mathcal{SN}$ denotes the Spiking Neuron Layer, $\op{BN}$ denotes Batch Normalization and $\bm{Q},\bm{K},\bm{V} \in \mathbb{R}^{T \times N\times D}$. Inspired by vanilla Self-Attention~\cite{vaswani2017attention}, SSA adds a scaling factor $s$ to control the large value of the matrix multiplication result, defined as:
\begin{equation}
\begin{aligned}
&{\rm{\text{SSA}}}(\bm{Q},\bm{K},\bm{V})={\mathcal{SN}}\left({\bm{Q}}~{\bm{K}^{\rm{T}}}~\bm{V} * s\right),\\
&\bm{X}^{\prime} ={\mathcal{SN}}(\op{BN}(\op{Dense}({\rm{\text{SSA}}}(\bm{Q},\bm{K},\bm{V})))),
\label{eq:ssa}
\end{aligned}
\end{equation}
where $\bm{X}^{\prime}\in\mathbb{R}^{T\times N \times D}$ are the updated spike sequences. It should be noted that SSA operates independently at each time step. In practice, $T$ represents an independent dimension for the $\mathcal{SN}$ layer. In other layers, it is merged with the batch size. Based on Equation (\ref{eq:spikeqkv}), the spike sequences $\bm{Q}$ and $\bm{K}$ produced by the $\mathcal{SN}$ layers $\mathcal{SN}_Q$ and $\mathcal{SN}_K$, respectively, naturally have non-negative values ($0$ or $1$). Consequently, the resulting attention map is also non-negative. Therefore, according to Equation (\ref{eq:ssa}), there is no need for softmax normalization to ensure the non-negativity of the attention map, and direct multiplication of $\bm{Q}$, $\bm{K}$ and $\bm{V}$ can be performed. This approach significantly improves computational efficiency compared to vanilla Self-Attention.

However, it is essential to note that SSA remains an operation with a computational complexity of $\mathcal{O}(N^2)$.\footnote{Although SSA can be decomposed with an $\mathcal{O}(N)$ attention scaling, this complexity hides large constants, causing limited scalability in practical applications. For more detailed analysis, refer to \textbf{Time Complexity Analysis of FW vs. SSA} Section.}
Within the spike-form frameworks, we firmly believe that SSA is not essential, and there exist simpler sequence mixing mechanisms that can efficiently extract sparse visual features as alternatives.

\subsection{Fourier Transform}

The Fourier Transform (FT) decomposes a function into its constituent frequencies. For an input spike sequence $\bm{x} \in \mathbb{R}^{N \times D}$ at a specific time step in $\bm{X}$, we utilize the FT to transform information from different dimensions, including 1D-FT and 2D-FT.

The discrete 1D-FT along the sequence dimension $\mathcal{F}_{\rm{seq}}$ to extract sparse visual features is defined by the equation:
\begin{equation}
    \label{eq:1d-dft}
    \bm{x}^{\prime}_n =\mathcal{F}_{\rm{seq}}(\bm{x}_n) =\sum_{k=0}^{N-1} \bm{x}_k e^{-{\frac{2\pi i}{N}} k n}, n=0,...,N-1, 
\end{equation}
where $i$ represents the imaginary unit. For each value of $n$, the discrete 1D-FT generates a new representation $\bm{x}^{\prime}_n\in\mathbb{R}^D$ as a sum of all of the original input spike features $\bm{x}_n\in\mathbb{R}^D$.\footnote{It is important to note that the weights in Equation~(\ref{eq:1d-dft}) are fixed constant and can be pre-calculated for all spike sequences.} 

Similarly, the discrete 2D-FT along the feature and sequence dimensions $\mathcal{F}_{\rm{seq}}(\mathcal{F}_{\rm{f}})$ is defined by the equation:
\begin{equation}
    \label{eq:2d-dft}
    \bm{x}^{\prime}_n =\mathcal{F}_{\rm{seq}}(\mathcal{F}_{\rm{f}}(\bm{x}_n)), n=0,...,N-1.
\end{equation}
Notably, equations (\ref{eq:1d-dft}) and (\ref{eq:2d-dft}) only consider the real part of the result. Therefore, there is no need to modify the subsequent MLP sub-layer or output layer to handle complex numbers.

\subsection{Wavelet Transform}

Wavelet Transform (WT) is developed based on Fourier Transform to overcome the limitation of Fourier Transform in capturing local features in the spatial domain. 

The discrete 1D-WT along the sequence dimension $\mathcal{W}_{\rm{seq}}$ to extract sparse visual features is defined by the equation:
\begin{align}
    \label{eq:1d-dwt}
    &\bm{x}^{\prime}_{n} =\mathcal{W}_{\rm{seq}}(\bm{x}_{n})=\frac{1}{\sqrt{N}}\Big[\bm{T}_{\varphi}(0,0)*\varphi(\bm{x}_{n}) + \sum\limits_{j=0}^{J-1}\sum\limits_{k=0}^{2^j-1}\bm{T}_{\psi}(j,k)*\psi_{j,k}(\bm{x}_{n})\Big], \\
    &\bm{T}_{\varphi}(0,0) = \frac{1}{\sqrt{N}}\sum\limits_{k=0}^{N-1}\bm{x}_{k}*\varphi({\bm{x}_{k}}),\ \  \bm{T}_{\psi}(j,k)=\frac{1}{\sqrt{N}}\sum\limits_{k^{\prime}=0}^{N-1}\bm{x}_{k^{\prime}}*\psi_{j,k}({\bm{x}_{k^{\prime}}}),
\end{align}
where $n=0, ..., N-1$, $N=2^J$ ($N$ is typically a power of $2$), $*$ denotes element-wise multiplication, $\bm{T}_{\varphi}(0,0)$ are the approximation coefficients, $\bm{T}_{\psi}(j,k)$ are the detail coefficients, $\varphi(x)$ is the scaling function, and $\psi_{j,k}(x)=2^{j/2}\psi(2^jx-k)$ is the wavelet function. 
Here, we use the Haar scaling function and Haar wavelet function for example, which is defined by the equation:
\begin{equation}
\varphi(x)=\left\{
	\begin{aligned}
	&1 \quad 0\leq x<1\\
	&0 \quad \rm{otherwise}\\
	\end{aligned}
	\right.,\ \ \ \ 
 \psi(x)=\left\{
	\begin{aligned}
	&1 \quad 0\leq x<0.5\\
 	&-1 \quad 0.5\leq x<1\\
	&0 \quad \rm{otherwise}\\
	\end{aligned}
	\right.,
\end{equation}


Similarly, the discrete 2D-WT along the feature and sequence dimensions $\mathcal{W}_{\rm{seq}}(\mathcal{W}_{\rm{f}})$ is defined by the equation:
\begin{equation}
    \label{eq:2d-dwt}
    \bm{x}^{\prime}_n =\mathcal{W}_{\rm{seq}}(\mathcal{W}_{\rm{f}}(\bm{x}_n)), n=0,...,N-1, 
\end{equation}

In the subsequent experimental section, we also delve into the exploration of different basis functions as well as their potential combinations.

\section{Method}
Following a standard Vision Transformer architecture, the vanilla spikformer incorporates several key components, including the Spiking Patch Splitting (SPS) module, Spikformer Encoder Layers, and a Classification head for visual classification tasks. 
Here, we directly replace vanilla SSA head with the FW head to efficiently manage spike-form features.

\begin{figure*}[tb]
\centering
\includegraphics[width=0.95\textwidth]{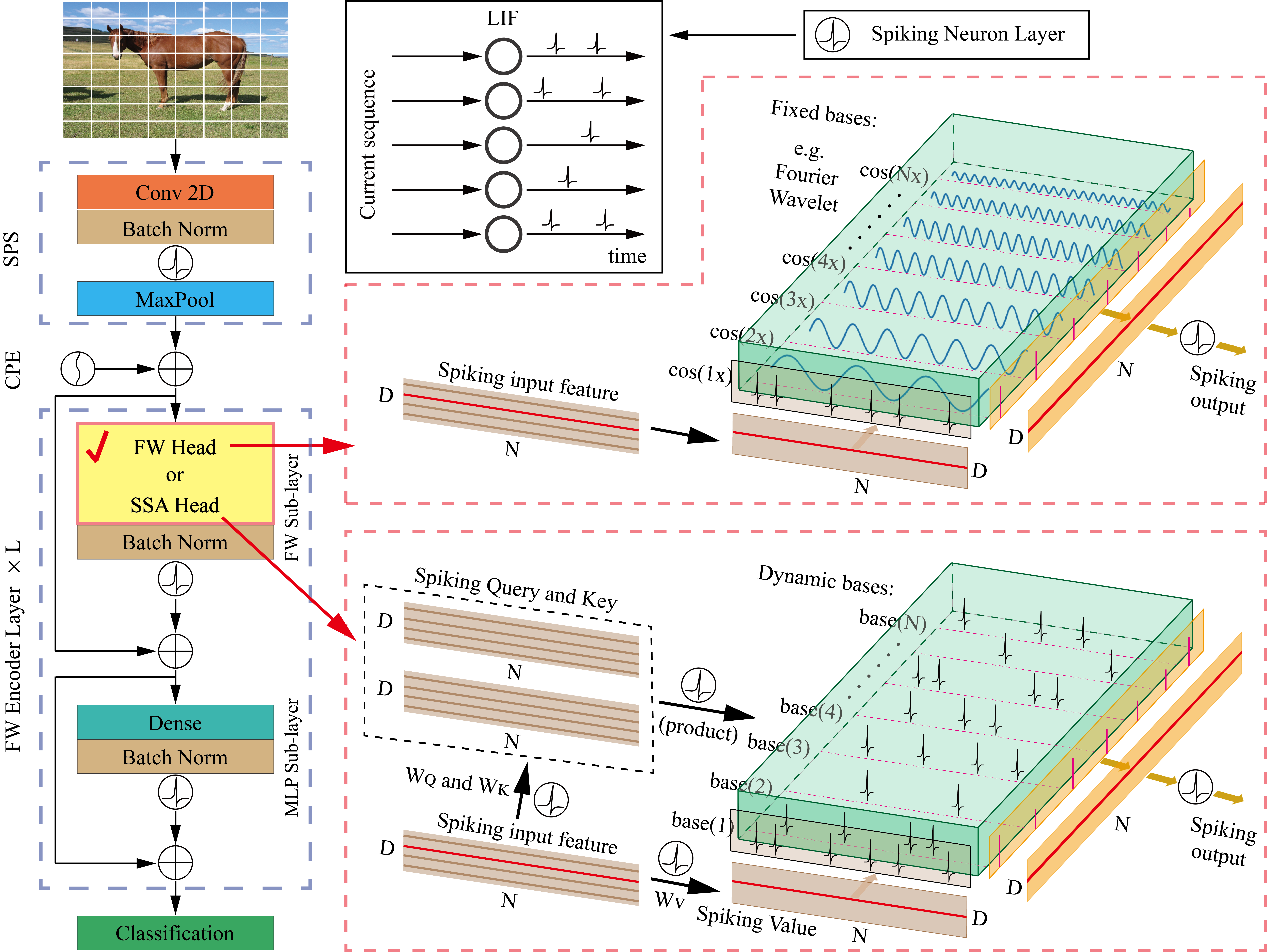}
\caption{The overall architecture of our proposed FWformer. It mainly consists of three components: (1) Spiking Patch Splitting (SPS) module, (2) FWformer Encoder Layer, and (3) Classification Layer. Additionally, we highlight the similarities between the FW head and SSA head at a single time step, which inspires us to choose the former as an exploration for more efficient calculations within the spike-form framework.}
\label{model}
\end{figure*}

In the following sections, we provide an overview of our proposed FWformer in Figure~\ref{model}, followed by a detailed explanation of the FW head. Finally, we compare the time complexity of both of these two heads.

\subsection{Overall Architecture}

We provide Figure~\ref{model} for an overview of our FWformer. First, for a given 2D image sequence $\bm{I}\in \mathbb{R}^{T \times C\times H\times W}$,\footnote{In the event-based video datasets, the data shape is $\bm{I} \in \mathbb{R}^{T \times C\times H\times W}$, where $T$, $C$, $H$, and $W$ denote the time step, channel, height, and width, respectively. In static datasets, a 2D image $\bm{I}_s \in \mathbb{R}^{C\times H\times W}$ needs to be repeated $T$ times to form an image sequence.} the goal of the Spiking Patch Splitting (SPS) module is to linearly project it into a $D$-dimensional spike-form feature and split this feature into a sequence of $N$ flattened spike-form patches $\bm{P}\in \mathbb{R}^{T \times N \times D}$.
Following the approach of the vanilla spikformer, the SPS module employs convolution operations to introduce inductive bias~\cite{xiao2021early}.

Second, to generate spike-form Relative Position Embedding (RPE), the Conditional Position Embedding (CPE) generator~\cite{chu2021twins} is utilized in the same manner as the spikformer. The RPE is then added to the patch sequence $\bm{P}$, resulting in $\bm{X}_0\in \mathbb{R}^{T \times N \times D}$.

Third, the $L$-layer FW Encoder is designed to manage $\bm{X}_0$. Different from spikformer encoder layer with SSA head, our FW Encoder Layer consists of an FW sub-layer and an MLP sub-layer, both with batch normalization and Spiking Neuron Layer. Residual connections are also applied to both the modules.
The FW head in FW sub-layer serves as a critical component in our encoder layer, providing an efficient method for spike-form sparse representation. 
We have provided two implementations for FW head, including Fourier Transform (FT) and Wavelet Transform (WT). Many works in the past have used FT and WT to alternate between the spatial and frequency domains, allowing for efficient analysis of signals. While in this paper we treat them as structured basis functions with prior knowledge for information transformation. These implementations will be thoroughly analyzed in the next section.

Finally, following the processing in spikformer, a Global Average-Pooling (GAP) operation is applied to the resulting spike features, generating a $D$-dimensional feature. The feature is then fed into the Classification module consisting of a spiking fully-connected (SFC) layer, which produces the prediction $\bm{Y}$. The formulation of our FWformer can be expressed as follows:
\begin{align}
\bm{P}&={\rm{SPS}}\left(\bm{I}\right),  \\
{\bm{\text{RPE}}}&={\rm{CPE}}(\bm{P}), \\
\bm{X}_0 &= \bm{P} + \bm{\text{RPE}}, \\
\bm{X}^{\prime}_l &= \mathcal{SN}({\rm{BN}}({\rm{FW}}(\bm{X}_{l-1}))) + \bm{X}_{l-1}, \\
 \bm{X}_l &= \mathcal{SN}({\rm{BN}}({\rm{MLP}}(\bm{X}^{\prime}_l))) + \bm{X}^{\prime}_l, \\
 \bm{Y} &= \op{SFC}(\op{GAP}(\bm{X}_L)),
\end{align}
where ${{\bm{I}}} \in \mathbb{R}^{T \times C\times H\times W}$, $\bm{P}\in \mathbb{R}^{T\times N\times D}$, ${\bm{\text{RPE}}}\in \mathbb{R}^{T \times N\times D}$, $\bm{X}_0\in \mathbb{R}^{T \times N\times D}$, $\bm{X}^{\prime}_l\in \mathbb{R}^{T \times N\times D}$, $\bm{X}_l\in \mathbb{R}^{T \times N\times D}$ and $l=1, ..., L$.

Moreover, the Membrane Shortcut (MS), which has been applied in many existing works~\cite{chen2023training, yao2024spike}, is also utilized in our model for comparison. It establishes a shortcut between the membrane potential of spiking neurons in various layers to enhance performance and increase biological plausibility~\cite{yao2024spike}.

\subsection{The FW head}\label{sec:lt}

Given input spike sequences $\bm{X} \in \mathbb{R}^{T \times N \times D}$, these features are then transformed into spiking sequences $\bm{X}^{\prime} \in \mathbb{R}^{T \times N \times D}$ through a $\mathcal{SN}$ layer. The formulation can be expressed as:
\begin{equation}
\begin{aligned}
&{\rm{FW}}(\bm{X})={\rm{FT}}(\bm{X}) ~or~ {\rm{WT}}(\bm{X}), \\
&\bm{X}^{\prime} ={\mathcal{SN}}(\op{BN}(({\rm{FW}}(\bm{X})))),
\label{eq:FW}
\end{aligned}
\end{equation}

In contrast to the SSA head in Equation~(\ref{eq:ssa}), the FW head does not involve any learnable parameters or Self-Attention calculations. Here, we can choose from Fourier Transform (FT) and Wavelet Transform (WT) with fixed basis functions. We can also combine different wavelet bases to form a superior function, which is defined as follows:

\begin{equation}
    \begin{gathered}
        Base = a \cdot Base1 + b \cdot Base2 + c \cdot Base3, \\
        \text{FW}(\bm{X}) = Base(\bm{X}),
    \end{gathered}
    \label{eq:combination}
\end{equation}

where $a$, $b$, $c$ are learnable parameters, and $Base1$, $Base2$, $Base3$ are selected bases. Since Wavelet Transform is a linear transformation, Equation~\ref{eq:combination} can also be written as:
\begin{equation}
{\rm{FW}}(\bm{X}) = a \cdot Base1(\bm{X}) + b \cdot Base2(\bm{X}) + c \cdot Base3(\bm{X})
\label{eq:final}
\end{equation}

Further analysis and experiments will be conducted based on the proposed FW head in the following sections.

\subsection{Time Complexity Analysis of FW vs. SSA}\label{complex}

We make a time complexity analysis between SSA, Fourier Transform (FT), and Wavelet Transform (WT). The results are presented in Table~\ref{tab:complex}. In the subsequent experimental section, we also conduct a more specific comparison of the training and inference speeds between FW and SSA under the same conditions.

\begin{wraptable}[]{t}{0.4\textwidth}
    \centering
    \caption{The time complexity for different methods. We have $N = 64$, $D = 384\ \text{or}\ 256$, and $d = 32$.}
    \resizebox{0.4\columnwidth}{!}{
    \begin{tabular}{c | c }
        \toprule
        Methods & Time Complexity \\
        \midrule
        SSA~\cite{zhou2022spikformer} & $\mathcal{O}(N^2d)$\ \rm{or}\ $\mathcal{O}(Nd^2)$\\
        1D-FFT & $\mathcal{O}(N\log N)$\\ 
        2D-FFT & $\mathcal{O}(D\log D)+\mathcal{O}(N\log N)$ \\ 
        2D-WT & $\mathcal{O}(D\log D)+\mathcal{O}(N\log N)$ \\ \bottomrule
    \end{tabular} 
    }
        \label{tab:complex}
\end{wraptable}

In SSA (Equation (\ref{eq:ssa})), since there is no softmax operation, the order of calculation between $\bm{Q}$, $\bm{K}$, and $\bm{V}$ can be changed: either $\bm{QK}^{\rm{T}}$ followed by $\bm{V}$, or $\bm{K}^{\rm{T}}\bm{V}$ followed by $\bm{Q}$. The former has a time complexity of $\mathcal{O}(N^2d)$, while the latter has $\mathcal{O}(Nd^2)$, where $d$ is the feature dimension per head.\footnote{In practice, the SSA in Equation (\ref{eq:ssa}) can be extended to multi-head SSA. In this case, $d = D/H$, where $H$ is the number of heads.} The second complexity, $\mathcal{O}(Nd^2)$, cannot be simply considered as $\mathcal{O}(N)$ due to the large constant $d^2$ involved. 
Only when the sequence length $N$ is significantly larger than the feature dimension per head $d$ does it demonstrate a significant computational efficiency advantage over the first complexity, $\mathcal{O}(N^2d)$.

In our implementation, we utilize the Fast Fourier Transform (FFT) algorithm to compute the discrete FT.
Specifically, we employ the Cooley-Tukey algorithm~\cite{cooley1965algorithm}, which recursively expresses the discrete FT of a sequence of length $N=N_1N_2$ in terms of $N_1$ smaller discrete FTs of size $N_2$, reducing the time complexity to $\mathcal{O}(N\log N)$ for discrete 1D-FT along the sequence dimension. Similarly, for discrete 2D-FT first along the feature dimension and then along the sequence dimension, the time complexity is $\mathcal{O}(D\log D)+\mathcal{O}(N\log N)$. In general, the complexity of WT is comparable to that of FFT~\cite{gonzales1987digital}.

\section{Experiments}

We conduct experiments on event-based video datasets (CIFAR10-DVS and DvsGesture), as well as static image datasets (CIFAR10 and CIFAR100). The FWformer is trained from scratch and compared with existing methods, including spikformer with SSA and its variant. More analyses are also given about the effects of different wavelet bases and their combinations.

\subsection{Experiment Settings}
To ensure a fair comparison, we ensure the same configurations of spikformer with SSA for datasets, implementation details, and evaluation metrics. To conduct the experiments, we implement the models using PyTorch and SpikingJelly~\cite{SpikingJelly}\footnote{\url{https://github.com/fangwei123456/spikingjelly}}.
All experiments are conducted on NVIDIA A100 GPU.
\paragraph{Event-based Video Datasets} 
For the CIFAR10-DVS and DvsGesture datasets, which have an image size of $128 \times 128$, we employ the Spiking Patch Splitting (SPS) module with a patch size of $16 \times 16$. This configuration splits each image into a sequence with a length $N$ of $64$ and a feature dimension $D$ of $256$.
We utilize $2$ FWformer encoder layers and set the time step of the spiking neuron to $16$. The training process consists of $106$ epochs for CIFAR10-DVS and $200$ epochs for DvsGesture. We employ the AdamW optimizer with a batch size of $16$. The learning rate is initialized to $0.1$ and reduced using cosine decay. Additionally, data augmentation techniques, as described in~\cite{li2022neuromorphic}, are applied specifically to the CIFAR10-DVS dataset.
\paragraph{Static Image Datasets}
For the CIFAR10/100 datasets featuring an image size of $32 \times 32$, we employ the SPS module with a patch size of $4 \times 4$, which splits each image into a sequence of length $N=64$ and a feature dimension of $D=384$. For the FWformer Encoder, we use $4$ layers, and the time-step of the spiking neuron is set to $4$.  During training, we utilize the AdamW optimizer with a batch size of $128$. The training process spans $400$ epochs, with a cosine-decay learning rate starting at $0.0005$. Following the approach outlined in~\cite{yuan2021tokens}, we apply standard data augmentation techniques such as random augmentation, mixup, and cutmix during training.

\subsection{Accuracy Performance}
We evaluate the accuracy performance on visual classification tasks, utilizing Top-1 accuracy (Top-1 acc.) as the performance metric. The results of our FWformer, spikformer with SSA, and other existing methods (both SNNs and ANNs, including the spikformer variant~\cite{yao2024spike}) on event-based video datasets as well as static image datasets are presented in Table~\ref{tab:acc}.

\begin{table}[htbp]
\centering
\caption{Accuracy performance comparison of our method with existing methods on CIFAR10-DVS (DVS10), DvsGesture (DVS128), CIFAR10, and CIFAR100. Our FWformer ($*$ means we replace vanilla residual connection with Membrane Shortcut (MS)) outperforms spikformer with SSA on event-based video datasets in terms of Top-1 acc. and achieves comparable accuracy on static datasets (the text in \textbf{bold} indicates the best results). It is necessary to mention that 2-256 signifies a configuration with 2 encoder layers and a feature dimension of 256.}
\begin{minipage}{0.48\textwidth}
\centering
\resizebox{1\columnwidth}{!}{
    \begin{tabular}{ccccc}
    \hline
    \toprule
    Methods & Architecture &  \makecell{Time Step\\(DVS10\textbf{/}128)} & \makecell{Top-1 acc.\\(DVS10\textbf{/}128)} \\
   \midrule
   LIAF~\cite{wu2021liaf} & LIAF-Net & 10\textbf{/}60 & 70.4\textbf{/}97.6\\
    TA-SNN~\cite{yao2021temporal} & TA-SNN & 10\textbf{/}60 & 72.0\textbf{/}98.6\\
     Rollout~\cite{kugele2020efficient} & -- & 48\textbf{/}240 & 66.8\textbf{/}97.2\\
    DECOLLE~\cite{kaiser2020synaptic} & -- &  -- \textbf{/}500 & -- \textbf{/}95.5\\
    tdBN~\cite{zheng2021going} & ResNet-19 &  10\textbf{/}40 & 67.8\textbf{/}96.9\\
    PLIF~\cite{fang2021incorporating} & --  & 20\textbf{/}20 & 74.8\textbf{/}97.6\\
    D-ResNet~\cite{fang2021deep} & Wide-7B-Net  & 16\textbf{/}16 & 74.4\textbf{/}97.9\\
    Dspike~\cite{li2021differentiable} & --  & 10\textbf{/} -- & $\rm{75.4}$\textbf{/}--\\
    SALT~\cite{kim2021optimizing} & --  & 20\textbf{/} -- & 67.1\textbf{/}--\\
    DSR~\cite{meng2022training} & -- & 10\textbf{/} -- & $\rm{77.3}$\textbf{/}--\\
    \midrule

    {SDSA~\cite{yao2024spike}}
    & {Spikformer-2-256} & 16\textbf{/}16 & $\rm{80.0}$\textbf{/}{99.3}\\
    
    {SSA~\cite{zhou2022spikformer}}
    & {Spikformer-2-256} & 16\textbf{/}16 & $\rm{80.9}$\textbf{/}{98.3}\\
    \midrule
      \textbf{1D-FFT} & {FWformer-2-256} & 16\textbf{/}16 &  $\rm{80.5}$\textbf{/}{99.0}\\
      \textbf{2D-FFT} & {FWformer-2-256} & 16\textbf{/}16 &  $\rm{80.6}$\textbf{/}{98.4}\\
      \textbf{2D-WT-Haar} & {FWformer-2-256} & 16\textbf{/}16 &  $\rm{81.0}$\textbf{/}{98.5}\\

      \textbf{1D-FFT}* & {FWformer-2-256} & 16\textbf{/}16 &  $\rm{80.8}$\textbf{/}{\textbf{99.5}}\\
      \textbf{2D-FFT}* & {FWformer-2-256} & 16\textbf{/}16 &  $\rm{80.7}$\textbf{/}{98.2}\\
      \textbf{2D-WT-Haar}* & {FWformer-2-256} & 16\textbf{/}16 &  $\rm{\textbf{81.2}}$\textbf{/}{99.1}\\
    \bottomrule
    \hline
\end{tabular}
}
\end{minipage}
\hspace{0.2cm}
\begin{minipage}{0.48\textwidth}
\centering
\resizebox{1\columnwidth}{!}{
    \begin{tabular}{ccccc}
    \hline
    \toprule
    Methods & Architecture& \makecell{Time\\Step} &  \makecell{Top-1 acc.\\(CIFAR10\textbf{/}100)} \\
   \midrule
    Hybrid training~\cite{rathienabling} & VGG-11 &125 & 92.22\textbf{/}67.87\\
    Diet-SNN~\cite{rathi2020diet} &ResNet-20 &10\textbf{/}5 & 92.54\textbf{/}64.07\\
    STBP~\cite{wu2018spatio} &CIFARNet &12& 89.83\textbf{/}--\\
   STBP NeuNorm~\cite{wu2019direct} &CIFARNet &12&90.53\textbf{/}--\\
   Dspike~\cite{li2021differentiable} & -- & 6 &  94.3\textbf{/}74.2\\
    TSSL-BP~\cite{zhang2020temporal} &CIFARNet &5 & 91.41\textbf{/}--\\
    STBP-tdBN~\cite{zheng2021going} &ResNet-19 & 4 &  92.92\textbf{/}70.86\\
   TET~\cite{dengtemporal}  &ResNet-19  & 4 &  94.44\textbf{/}74.47\\
   \midrule
   \multicolumn{1}{c}{\multirow{2}{*}{{ANN Methods}}}  & ResNet-19 &1 &   94.97\textbf{/}75.35\\
     & {Transformer-4-384}  &1 &  \textbf{96.73}\textbf{/}\textbf{81.02} \\
     \midrule

     SDSA~\cite{yao2024spike}
    & {Spikformer-4-384}& 4 & \textbf{95.6}\textbf{/}\textbf{78.4}\\
    
    {SSA~\cite{zhou2022spikformer}}
    & {Spikformer-4-384}& 4 & 95.51\textbf{/}78.21 \\
    
    \midrule
  \textbf{1D-FFT} & {FWformer-4-384}& 4 &  {94.9}\textbf{/}{77.3}\\
    \textbf{2D-FFT}  & {FWformer-4-384}& 4 &  {95.1}\textbf{/}{77.9}\\
\textbf{2D-WT-Haar}  & {FWformer-4-384}& 4 &  {95.2}\textbf{/}{78.1}\\

\textbf{1D-FFT}* & {FWformer-4-384}& 4 &  {95.5}\textbf{/}{78.0}\\
    \textbf{2D-FFT}*  & {FWformer-4-384}& 4 &  {95.0}\textbf{/}{78.3}\\
\textbf{2D-WT-Haar}*  & {FWformer-4-384}& 4 &  \textbf{95.6}\textbf{/}{78.2}\\
    \bottomrule
    \hline
    \end{tabular}
}
    \end{minipage}
  \label{tab:acc}
\end{table}

Our FWformer achieves remarkable accuracy, reaching $81.2\%$ on CIFAR10-DVS with 2D-WT-Haar, and an impressive $99.5\%$ on DvsGesture with 1D-FFT. The performances surpass the spikformer with SSA by $0.3\%$ and $1.2\%$, respectively. While on static datasets, our FWformer variants demonstrate comparable Top-1 accuracy. The results demonstrate the advantage of our methods, particularly on event-based video datasets.

\subsection{Computational Costs and Speed Performance}\label{results}

Furthermore, we conduct a comprehensive comparison between existing works and our FWformer in terms of GPU memory usage, training speed, and inference speed, ensuring identical operating conditions. The training speed represents the time taken for the forward and back-propagation of a batch of data, while the inference speed denotes the time taken for the forward-propagation of a batch of data in milliseconds (ms). To minimize variance, we calculate the average time spent over 100 batches. The results are presented in Table~\ref{tab:mem} (DWT-C means 2D-WT combination with learnable parameters, which will be discussed in the next subsection). We also provide an analysis of energy efficiency in \textbf{Appendix}.

In the case of event-based video datasets, our FWformer achieves a significant reduction in the number of parameters, approximately $20\%$, under identical hyperparameter configurations and operating conditions. This reduction, attributed to the absence of learnable parameters, translates to around $4\%$-$5\%$ memory savings. Moreover, our FWformer demonstrates remarkable improvements in both training and inference speeds, showing increases of approximately $9\%$-$51\%$ and $33\%$-$70\%$, respectively, compared to SSA. While in the case of static datasets, our FWformer also shows several advantages under identical hyperparameter configurations and operating conditions. It achieves a notable reduction in the number of parameters, approximately $25\%$, leading to memory savings of around $26\%$. Furthermore, our FWformer enhances both training and inference speeds by approximately $18\%$-$29\%$ and $19\%$-$61\%$, respectively.

\begin{table}[htbp]
\centering
\caption{Memory usage and speed performance comparison of our method with existing methods on CIFAR10-DVS (DVS10), DvsGesture (DVS128) and CIFAR-static (CIFAR10 and CIFAR100). Our FWformer outperforms spikformer with SSA~\cite{zhou2022spikformer} and its variant~\cite{yao2024spike} when comparing GPU memory usage, training speed and inference speed under identical operating conditions.}
  
\begin{minipage}{0.48\textwidth}
\centering
\resizebox{1.0\columnwidth}{!}{
    \begin{tabular}{ccccc}
    \hline
    \toprule
    Methods & \makecell{Param\\(M)} & \makecell{Memory\\(DVS10\textbf{/}128)\\(GB)}  & \makecell{Training Speed\\(DVS10\textbf{/}128)\\(ms/batch)}& \makecell{Inference Speed\\(DVS10\textbf{/}128)\\(ms/batch)}  \\
   \midrule
   STBP-tdBN~\cite{zheng2021going} & 12.63 & 25.86\textbf{/}25.87 & 65\textbf{/}194  & 27\textbf{/}98 \\
      TET~\cite{dengtemporal} & 12.63 & 36.13\textbf{/}36.17 & 71\textbf{/}203 & 22\textbf{/}77 \\

      {SDSA~\cite{yao2024spike}} & 2.59 & 9.02\textbf{/}9.03 & 73\textbf{/}245  & 29\textbf{/}101 \\

    {SSA~\cite{zhou2022spikformer}} & 2.59 & 9.02\textbf{/}9.03 & 76\textbf{/}246  & 30\textbf{/}105 \\
    \midrule
                                                  
      \textbf{1D-FFT}   &\textbf{2.06}& 8.67\textbf{/}\textbf{8.71} & \textbf{51}\textbf{/}\textbf{121} & \textbf{11}\textbf{/}\textbf{32} \\
      \textbf{2D-FFT}   &\textbf{2.06}& \textbf{8.54}\textbf{/}8.74 & 55\textbf{/}135  & 21\textbf{/}37 \\
      \textbf{2D-WT-Haar}  &\textbf{2.06}& 8.70\textbf{/}8.73 & 62\textbf{/}139  & 21\textbf{/}46 \\

\textbf{DWT-C}  &\textbf{2.06}&  8.55\textbf{/}8.74 &  69\textbf{/}158  &  23\textbf{/}48 \\

    \bottomrule
    \hline
    \end{tabular}
    }
\end{minipage}
\hspace{0.15cm}
\begin{minipage}{0.48\textwidth}
\centering
    \resizebox{1\columnwidth}{!}{
    \begin{tabular}{ccccc}
    \hline
    \toprule
    Methods & \makecell{Param\\(M)}& \makecell{Memory\\(CIFAR-static)\\(GB)}  & \makecell{Training Speed\\(CIFAR-static)\\(ms/batch)}& \makecell{Inference Speed\\(CIFAR-static)\\(ms/batch)} \\
   \midrule
    STBP-tdBN~\cite{zheng2021going} & 12.63 & 8.02 & 155 & 20 \\
      TET~\cite{dengtemporal} & 12.63 & 8.19 & 148 & 23 \\

    {SDSA~\cite{yao2024spike}} 

    & 9.32 & 11.69 & 162 & 33 \\
    
    {SSA~\cite{zhou2022spikformer}} 

    & 9.32 & 11.69 & 166 & 31 \\
    \midrule

  \textbf{1D-FFT}  &\textbf{6.96}    &\textbf{8.61}&\textbf{118}&\textbf{12}\\
    \textbf{2D-FFT}   &\textbf{6.96}    &8.75&122&13\\
\textbf{2D-WT-Haar}   &\textbf{6.96}    &9.33&121&19\\

\textbf{DWT-C}   &\textbf{6.96}    &9.86&136&25\\

    \bottomrule
    \hline
    \end{tabular}%
    }
    \end{minipage}%
  \label{tab:mem}%
\end{table}%

\subsection{From orthogonal to non-orthogonal bases}\label{more_analysis}

In the previous experiments, the Haar base was used as the default choice for Wavelet Transform. We have also compared the performance of some other wavelet bases including Db1, Bior1.1, and Rbio1.1, each having different functions for deconstructing the spike-form feature while maintaining orthogonality. The results on CIFAR10-DVS and DvsGesture are presented in Table~\ref{tab:different_base}. Interestingly, most alternative basis functions yield similar Top-1 accuracy. Their performance is comparable to or even better than that of spikformer.
It is essential to highlight that Wavelet Transform offers numerous different basis function options, and our exploration has not been exhaustive. Investigating the influence of more basis function choices on accuracy, as well as the possibility of identifying superior basis functions, is an avenue for future research.

\begin{figure}[tb]
\centering
\includegraphics[width=0.95\textwidth]{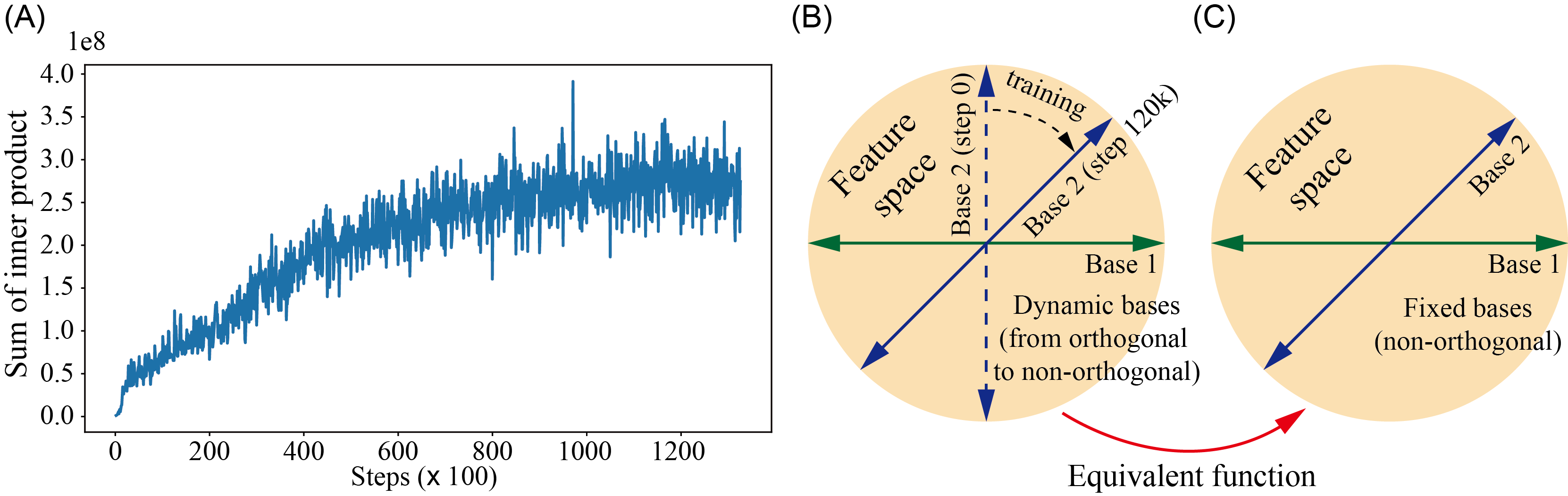}
\caption{(A) We treat spiking self-attention as a set of basis functions and proceed to measure the changes in their orthogonality throughout the training process. (B) A diagram visualizing how the basis functions, spanning a feature space, are transformed from orthogonal to non-orthogonal, with only two axes used for simplification. (C) A diagram visualizing our endeavor to employ fixed non-orthogonal bases.}
\label{base}

\end{figure}

However, a more interesting question arises: Is it always necessary to pursue orthogonality? Although in many cases, orthogonality signifies sparse and efficient information representation, neural networks may show the opposite phenomenon in the actual training process, that is, parameters naturally tend toward overlapping representations, and SSA is no exception. To illustrate this, we treat SSA as basis functions as proposed in the previous sections, and then quantitatively measure changes in their orthogonality during training. We calculate the inner product of each row vector (one base) in $Q \times K$ with others (other bases) and sum them in each training step. The variation trend is shown in Figure~\ref{base} (A). Initially, network parameters are nearly orthogonal at initialization, but their orthogonality is continuously decreasing during training. A diagram visualizing how the basis functions change during training is provided in Figure~\ref{base} (B). Inspired by this phenomenon, we further explore the combination of different wavelet bases to form fixed non-orthogonal basis functions, as depicted in Figure~\ref{base} (C). We assume that the pre-trained dynamic bases may serve an equivalent function to the fixed non-orthogonal bases.

Here we choose Bior1.1, Haar, and Db1 for further exploration. To search for the proper coefficients of their combinations, we set an initial set of learnable coefficients first and then conduct training. The new FW head will have three parameters to learn, which exert minimal influence on the overall computation of the network but play a crucial role in finding suitable combinations. The results are presented in Table~\ref{tab:different_base}. Consistent with our hypothesis, the use of fixed non-orthogonal basis functions further improves accuracy performance ($0.4\%$-$1.5\%$ improvements on event-based video datasets, compared to vanilla spikformer).


\begin{wraptable}[]{r}{0.4\textwidth}
\caption{Accuracy performance of different wavelet bases as well as their combination with learnable parameters (DWT-C) on DVS10 and DVS128 datasets ($*$ means replacing vanilla residual connection with Membrane Shortcut (MS)).}
    \centering
    \resizebox{0.4\columnwidth}{!}{
    \begin{tabular}{ccc}
    \toprule
    Methods & Architecture  & \makecell{Top-1 acc.\\DVS10/128} \\
    \midrule
    Db1 & {FWformer-2-256} &  81.0\textbf{/}{98.7}\\
Bior1.1  & {FWformer-2-256} &  80.9\textbf{/}{98.2}\\
Rbio1.1  & {FWformer-2-256} &  80.4\textbf{/}{98.1}\\
    \midrule
 DWT-C & {FWformer-2-256} &  {\textbf{81.3}}\textbf{/}{99.1}\\
 {DWT-C}* & {FWformer-2-256} &  {81.2}\textbf{/}{\textbf{99.8}}\\
    \bottomrule
    \end{tabular}%
    }
    \label{tab:different_base}%

    \end{wraptable}%

We attempt to conduct a preliminary analysis of the situations in which our FWformer is applicable: In contrast to conventional signal processing, complex tasks such as NLP and ASR need the designed models to learn diverse syntactic and semantic relationships, which can hardly be represented simply by fixed basis functions such as Fourier bases. For this reason, the network has to form dynamic higher-order basis functions, which are adjusted by not only the changing inputs but also the parameter learning of the network itself. We can regard these basis functions in networks as hyper-parameters that need continuous adjustment. However, this also means each time the basis functions change, the rest of the network has to adapt accordingly, which is understandable in complex tasks but not necessary in some other cases (e.g. event-based video tasks). Moreover, within spike-form frameworks, the features are represented by such sparse spiking signals that the correlation between them is too weak to form closed similarity, so it is more suitable to use structured fixed basis functions (e.g. Fourier bases and Wavelet bases) containing prior knowledge to get a simplified network.

\section{Conclusion}

We present the FWformer that replaces SSA with spike-form FW head, based on the hypothesis that both of them use dynamic or fixed bases to transform information. The proposed model achieves comparable or better accuracy, higher training and inference speed, and reduced computational cost, on both event-based video datasets and static datasets. We analyze the orthogonality in SSA during training and assume that the pre-trained dynamic bases serve an equivalent function to the fixed bases, which inspires us to explore non-orthogonal combined bases and get even higher accuracy. Additionally, we provide an analysis of why and under what scenarios our FWformer is effective, indicating the promising refinement of new Transformers in the future, which is inspired by biological discovery and information theory.

\section{Appendix}

We estimate the theoretical energy consumption of FWformer mainly according to~\cite{yao2024spike, horowitz20141, zhou2022spikformer}. It is calculated by the following two equations:
\begin{align}
    \label{SOP}
    \text{SOPs}(l) = Rate \times T \times \text{FLOPs}(l),
\end{align}
\begin{equation}
\begin{aligned}
    \label{all}
    E_{FWformer} &= E_{MAC} \times \text{EL}_{Conv}^1 \\ 
    &+ E_{AC} \times (\sum_{k=2}^{K} \text{SOP}_{Conv}^k + \sum_{m=1}^{M} \text{SOP}_{FC}^m + \sum_{n=1}^{N} \text{SOP}_{FW}^n),
\end{aligned}
\end{equation}

$\text{SOPs}(l)$ means synaptic operations (the number of spike-based accumulate (AC) operations) of layer $l$, $Rate$ is the average firing rate of input spike train to layer $l$, $T$ is the time window of LIF neurons, and $\text{FLOPs}(l)$ refers to the floating point operations (the number of multiply-and-accumulate (MAC) operations) of layer $l$. We assume that the MAC and AC operations are implemented on the 45nm hardware~\cite{horowitz20141}, with $E_{MAC}=4.6pJ$ and $E_{AC}=0.9pJ$.

$\text{EL}_{Conv}^1$ represents the FLOPs of convolution module in ANNs. It is used for the first layer to convert static images into spike trains, which can also be written as $\text{SOP}_{Conv}^1$ for event-based video datasets, and $\text{SOP}_{Net}^{l}$ ($\text{SOP}_{Conv}^{k}$, $\text{SOP}_{FC}^{m}$, $\text{SOP}_{FW}^{n}$) is for the rest of FWformer.

The experimental settings are the same as in the main text. Each cell in the table below contains results presented in the form of OPs(G)\textbf{/}Power(mJ), where OPs refers to the total SOPs in a SNN model, and Power refers to the average theoretical energy consumption when predicting one sample from the datasets.

The results indicate that our methods can achieve a reduction in energy consumption of approximately 20$\%$$-$25$\%$ compared to SSA~\cite{zhou2022spikformer}, and 4$\%$$-$9$\%$ compared to its variant~\cite{yao2024spike}. This is primarily due to lower computational complexity of the FW head, as reflected in fewer total SOPs (OPs). Our FWformer demonstrates enhanced energy efficiency.

\begin{table}[htbp]
\centering
\caption{The theoretical energy consumption on DVS10 (CIFAR10-DVS), DVS128 (DvsGesture), CIFAR10 and CIFAR100 dataset. DWT-C means 2D-WT combination with learnable parameters.}
  
\begin{minipage}{0.48\textwidth}
\centering
\resizebox{1.0\columnwidth}{!}{
    \begin{tabular}{ccc}
    \hline
    \toprule
    Methods & \makecell{DVS10\\OPs(G)/Power(mJ)} & \makecell{DVS128\\OPs(G)/Power(mJ)} \\
   \midrule
    {SDSA~\cite{yao2024spike}} 
    & 1.561\textbf{/}0.816 & 1.620\textbf{/}0.713  \\
    {SSA~\cite{zhou2022spikformer}}
    & 1.852\textbf{/}0.943 & 1.914\textbf{/}0.822  \\
    \midrule
      1D-FFT  & 1.547\textbf{/}0.752 & 1.608\textbf{/}0.650 \\
      2D-FFT  & 1.548\textbf{/}0.752 & 1.609\textbf{/}0.653 \\
      2D-WT  & 1.549\textbf{/}0.753 & 1.609\textbf{/}0.651  \\
      DWT-C  & 1.553\textbf{/}0.753 & 1.613\textbf{/}0.652  \\
    \bottomrule
    \hline
    \end{tabular}
    }
\end{minipage}
\hspace{0.15cm}
\begin{minipage}{0.48\textwidth}
\centering
    \resizebox{1\columnwidth}{!}{
    \begin{tabular}{ccc}
    \hline
    \toprule
    Methods & \makecell{CIFAR10\\OPs(G)/Power(mJ)} & \makecell{CIFAR100\\OPs(G)/Power(mJ)} \\
   \midrule
   
    {SDSA~\cite{yao2024spike}}                            
    & 0.951\textbf{/}0.415 & 1.446\textbf{/}0.609  \\
    {SSA~\cite{zhou2022spikformer}}                       
   &  1.186\textbf{/}0.523  & 1.737\textbf{/}0.748 \\
    \midrule
      1D-FFT  & 0.942\textbf{/}0.392 & 1.438\textbf{/}0.578 \\
      2D-FFT  & 0.943\textbf{/}0.393  & 1.438\textbf{/}0.584 \\
      2D-WT  & 0.944\textbf{/}0.393  & 1.439\textbf{/}0.584 \\
      DWT-C  & 0.947\textbf{/}0.394  & 1.442\textbf{/}0.586 \\
    \bottomrule
    \hline
    \end{tabular}%
    }
    \end{minipage}%
  \label{tab:cost}%
\end{table}%

%
%
\bibliographystyle{splncs04}
\bibliography{main}

\end{document}